\documentclass{article}

\usepackage{PRIMEarxiv}

\usepackage{graphicx}%
\usepackage{multirow}%
\usepackage{amsmath,amssymb,amsfonts}%
\usepackage{amsthm}%
\usepackage{mathrsfs}%
\usepackage[title]{appendix}%
\usepackage{xcolor}%
\usepackage{textcomp}%
\usepackage{manyfoot}%
\usepackage{booktabs}%
\usepackage{algorithm}%
\usepackage{algorithmicx}%
\usepackage{algpseudocode}%
\usepackage{listings}%
\usepackage{soul}

\usepackage{amsfonts}
\usepackage{siunitx}
\usepackage{caption}
\usepackage{subcaption}
\usepackage{graphicx}
\usepackage{multirow}
\usepackage{hyperref}
\usepackage{tikz}
\usetikzlibrary{arrows.meta, positioning, shapes.geometric}

\title{Towards Real-Time Autonomous Navigation: Transformer-Based Catheter Tip Tracking in Fluoroscopy
\thanks{\textit{\underline{Citation}}: 
\textbf{Robertshaw, H., Hao, Y., Deng, W. et al. Toward real-time autonomous navigation: transformer-based catheter tip tracking in fluoroscopy. Int J CARS (2026). \url{https://doi.org/10.1007/s11548-026-03647-7}}} 
}

\author{
 Harry Robertshaw\thanks{Equal contribution} , Yanghe Hao\textsuperscript{\(\dag \)}, Weiyuan Deng, Benjamin Jackson, Nikola Fischer, Tom Vercauteren, \\ \textbf{Alejandro Granados, Thomas C. Booth\thanks{Corresponding author: \texttt{thomas.booth@kcl.ac.uk}}} \\
  Surgical \& Interventional Engineering \\
  School of Biomedical Engineering \& Imaging Sciences \\
  Kings College London \\
  London\\
   \And
 S.M.Hadi Sadati \\
  School of Engineering \& Materials Science\\
  Queen Mary London\\
  London \\
}

\begin{document}
\maketitle
\begin{abstract}
    \textbf{Purpose:} Mechanical thrombectomy (MT) improves stroke outcomes, but is limited by a lack of local treatment access. Widespread distribution of reinforcement learning (RL)-based robotic systems can be used to alleviate this challenge through autonomous navigation, but current RL methods require live device tip coordinate tracking to function. This paper aims to develop and evaluate a real-time catheter tip tracking pipeline under fluoroscopy, addressing challenges such as low contrast, noise, and device occlusion. 
    
    \textbf{Methods:} A multi-threaded pipeline was designed, incorporating frame reading, preprocessing, inference, and post-processing. Deep learning segmentation models, including U-Net, U-Net+Transformer, and SegFormer, were trained and benchmarked using two-class and three-class formulations. Post-processing involved two-step component filtering, one-pixel medial skeletonization, and greedy arc-length path following with contour fall-back. 
    
    \textbf{Results:} On manually-labeled moderate complexity fluoroscopic video data, the two-class SegFormer achieved a mean absolute error of 4.44\,mm, outperforming U-Net (4.60\,mm), U-Net+Transformer (6.20\,mm) and all three-class models (5.19-7.74\,mm). On segmentation benchmarks, the system exceeded state-of-the-art CathAction results with improvements of up to +5\% in Dice scores for three-segmentation. 
    
    \textbf{Conclusion:} The results demonstrate that the proposed multi-threaded tracking framework maintains stable performance under challenging imaging conditions, outperforming prior benchmarks, while providing a reliable and efficient foundation for RL-based autonomous MT navigation.
\end{abstract}

\keywords{catheter tip tracking \and fluoroscopy \and mechanical thrombectomy \and deep learning segmentation}

\section{Introduction}

    Mechanical thrombectomy (MT) is an essential endovascular intervention in treating stroke due to large vessel occlusion~\cite{Jadhav2021}. However, its efficacy declines beyond 7.3\,h of stroke onset, contributing to low treatment rates despite up to 15\% of patients being eligible~\cite{Saver2016,SSNAP2024}. MT can be associated with vessel perforations (1\%), procedure-related vessel dissections (2\%), or distal embolization of thrombus (9\%)~\cite{Berkhemer2015}. Cumulative X-ray exposure endangers clinicians with cancer and cataract risks, compounded by the musculoskeletal burden of lead garments~\cite{Klein2009,Madder2017}. Endovascular robotic platforms mitigate radiation exposure and ergonomic strain~\cite{Crinnion2022}, but their controller-operator design incurs high cognitive load and is susceptible to human error~\cite{Mofatteh2021}.

    Recent research has investigated AI-based autonomous MT navigation, using reinforcement learning (RL) and robotics~\cite{Robertshaw2024,Robertshaw1_2025}. The current state-of-the-art utilizes a `state-based' RL approach, requiring information such as the device tip coordinates, target coordinates, and previous actions~\cite{Robertshaw2023}. However, current methods for device tracking and tip localization are unproven under clinically realistic conditions, and it is unknown whether existing deep learning models can generalize across variable imaging, such as camera/fluoroscopy, while maintaining robust tip localization when it is hidden or poorly visible in real-time. To realize the benefits of previous autonomous endovascular navigation research, it is necessary to develop accurate and robust device tip localization methods, which can provide input to RL agents as they are transferred from \textit{in silico} to \textit{in vitro}, \textit{ex vivo}, and \textit{in vivo} testbeds. Furthermore, reliable tracking of both guidewire and catheter tip (i.e. 3-class formulation) will have a higher impact on training RL agents~\cite{Robertshaw2024} on environments related to real interventions. This will allow continuous development of autonomous tele-operated robotic platforms, promoting increased accessibility of expert-level treatment to underserved regions. 

    Current device tracking methods rely on deep networks such as U-Net and instance-segmentation models, aiming to deliver full-catheter masks at $\sim 8$\,fps with $< 1$\,mm tip error without manual setup~\cite{Ronneberger2015, Ambrosini2017}. Two-stage detectors with multi-task centerline/tip predictors and unsupervised optical-flow have been used to improve robustness when poorly visible due to low image contrast or when hidden by overlapping objects (occlusion)~\cite{Nguyen2020}. Additionally, `weakly supervised' learning has been shown to reduce annotation needs~\cite{Vlontzos2018}, while transformers have been used to capture broader context~\cite{Demoustier2023}. However, the absence of a large, open-source training dataset has limited performance when testing on complicated and realistic clinical environments. The CathAction dataset was recently introduced with $\sim 500{\small,}000$ annotated frames and $25{\small,}000$ masks with separate instrument classes, facilitating pixel-level tracking~\cite{Huang2024}. Segmentation results on this dataset are currently benchmarked using U-Net~\cite{Ronneberger2015}, TransUNet~\cite{Chen2024}, SwinUNet~\cite{Cao2023}, and SegViT~\cite{Zhang2022}. Most existing models have been evaluated primarily on \textit{in silico} or \textit{in vitro} phantom datasets, but it is unclear whether they generalize to \textit{in vitro} live-camera data used in experimental testbeds, or to real \textit{in vivo} clinical fluoroscopy. Furthermore, to our knowledge, the CathAction dataset has not yet been used to develop a live, real-time catheter tip detection or tracking algorithm across any experimental medium (\textit{in vitro} through to \textit{in vivo}), representing an important opportunity for advancing autonomous endovascular navigation.

    The aim of this study was to propose a deep learning-based catheter and guidewire segmentation and tip-tracking method, leveraging the CathAction dataset, to enable accurate, real-time device localization under clinically challenging conditions. The primary objective was to demonstrate that modern segmentation architectures (U-Net, U-Net+Transformer, and SegFormer) can be effectively integrated into a downstream tracking pipeline, rather than evaluated as standalone segmentation models, to achieve high accuracy and low tip-localization error. The secondary objectives were to evaluate the generalizability of the proposed tracking framework across clinically relevant footage, and to benchmark two-class (instrument vs.~background) versus three-class (catheter vs.~guidewire vs.~background) segmentation in the context of downstream tracking accuracy rather than segmentation performance alone. Our contributions are: 1) We introduce a novel guidewire and catheter tip-tracking pipeline based on skeletonization and multi-point sampling, improving localization in the presence of low image contrast, partial occlusion, and overlapping instruments; 2) we demonstrate that, when embedded within this tracking framework, modern segmentation architectures achieve improved performance compared to the existing benchmarks on the CathAction dataset, extending beyond prior segmentation-only evaluations; 3) We validate the proposed method across \textit{in vitro} phantom, \textit{in vivo} animal, and \textit{in vivo} clinical fluoroscopic data, demonstrating its feasibility for real-time catheter tip tracking and its potential role in enabling AI-driven endovascular navigation systems.

\section{Methods}

    \subsection{Dataset}\label{sec:dataset}

        \textbf{Training data:} The action-based dataset from CathAction was used for training, focusing on the segmentation task component~\cite{Huang2024}. The dataset comprised $25{\small,}000$ pixel-level reference standard masks for manually-segmented catheters and guidewires. These devices were captured in 569 fluoroscopic videos derived from two different sources: 1) five \textit{in vitro} human-like vascular silicone phantoms, offering homogeneous imaging conditions with consistent contrast and minimal motion artifacts; 2) five \textit{in vivo} porcine subjects, giving heterogeneous imaging conditions with cardiac motion, vessel deformation, and variable contrast agent distribution. As in the CathAction study, we use the phantom data for training. All videos were processed at $500 \times 500$ resolution and $24$\,fps.

        \noindent\textbf{Evaluation data:} Four types of data were used during evaluation: 1) \textit{CathAction}, models evaluated on a held-out porcine navigation videos from the CathAction dataset, which were not used during training; 2) \textit{in vitro offline RGB}, a $1{\small,}997$-frame RGB navigation video ($240 \times 320$ resolution) dataset using an in \textit{in vitro} patient vasculature phantom built from CT angiograms (with manual multi-class device tip labels); 3) \textit{in vitro live RGB}, a live RGB camera ($600 \times 400$ resolution) dataset showing navigation in the same \textit{in vitro} patient vasculature phantom. This dataset was used exclusively for qualitative evaluation and demonstration of real-time tracking performance and robustness. As no frame-level annotations were available and no fixed evaluation protocol was defined, quantitative metrics and frame counts were not computed for this dataset; 4) \textit{in vivo fluoroscopy}, an \textit{in vivo} human fluoroscopic dataset from five endovascular interventions (UK Research Ethics Committee 24/LO/0057) with manually-labeled multi-class ground truths with each consisting $\approx 900$ frames ($352 \times 640$ resolution). The \textit{in vivo fluoroscopy} dataset was stratified into three levels of complexity based on level of occlusion: Group 1 (high), 2 (moderate), and 3 (low). 

        All evaluation datasets outside CathAction are entirely independent from the training data and were used solely for assessing generalization across imaging modalities and conditions. The evaluation of our proposed model relies on publicly available and private datasets. The public dataset CatchAction~\cite{Jianu2024} is used to guarantee our results can be comparable against other approaches in the literature. In addition, patient-derived datasets were used under institutional data governance frameworks overseen with ethical approval. While these datasets cannot be released as open-access resources at present, they can be made available contingent upon further ethical approvals. However, all evaluation protocols, parameter settings, and implementation details are fully reported to support reproducibility on alternative datasets.
    
    \subsection{Network model}

        Deep-learning methods were used to achieve high segmentation accuracy while maintaining real-time tracking performance under the complex lighting and contrast conditions encountered in clinical practice~\cite{GHERARDINI2020105420}. We investigated three modern segmentation architectures, selected for their performance:

        \textbf{(1) Standard U-Net}~\cite{Ronneberger2015}: a standard three-stage U-Net in which the encoder comprises four convolutional blocks where each block performs two $3 \times 3$ convolutions (padding), batch normalization, ReLU, and $2 \times 2$ max-pooling, halving spatial resolution and doubling feature channels ($32 \rightarrow 64 \rightarrow 128 \rightarrow 256$). This is followed by a symmetrically structured 512-channel bottleneck block. The decoder then mirrors this design: each level begins with a $2 \times 2$ transposed convolution for upsampling (halving channels), concatenates the corresponding encoder feature map via skip connections, and applies two $3 \times 3$ conv$\rightarrow$BN$\rightarrow$ReLU layers ($256 \rightarrow 128 \rightarrow 64 \rightarrow 32$), before a final $1 \times 1$ convolution projects to three-class logits with per-pixel softmax. Note that we initialized this U-Net with 16 base feature channels and trained using the AdamW optimizer rather than CathAction's method of 64 channels and SGD. This modification was made to ensure architectural and optimization consistency across all tested backbones (U-Net, U-Net+Transformer, and SegFormer), and to meet real-time memory and latency constraints within the integrated tracking pipeline. Our goal was not to reproduce the absolute segmentation state-of-the-art configuration from CathAction, but to evaluate backbone behavior under a unified, deployment-oriented training protocol.
        
        \textbf{(2) U-Net+Transformer}: the U-Net backbone remains unchanged except that the 512-channel bottleneck is replaced by a transformer module: a convolutional patch embedding reshapes the $H \times W$ feature map into $D$-dimensional tokens, which, augmented with positional encodings, are processed by a 16-layer, 16-head self-attention Transformer (with $D \rightarrow 4D$ feed-forward). These tokens are then re-projected and reshaped into the original 512-channel feature map, from which the standard U-Net decoder yields the final segmentation.

        \textbf{(3) SegFormer}~\cite{xie2021segformer}: the architecture comprises a Mix Transformer B5 encoder backbone with a symmetric U-Net decoder. The encoder processes $512 \times 512 \times 3$ RGB inputs through hierarchical transformer blocks, extracting multi-scale features at progressively reduced spatial resolutions. The decoder path contains five levels with channel dimensions $256 \rightarrow 128 \rightarrow 64 \rightarrow 32 \rightarrow 16$: each level performs upsampling via transposed convolution and applies convolutional blocks with concurrent spatial and channel squeeze \& excitation (scSE) attention modules. A $1 \times 1$ convolution projects the 16-channel features to two-class logits, using a per-pixel softmax for binary segmentation.

    \subsection{Tracking Algorithm}

\begin{figure}[ht]
            \centering
            \includegraphics[width=0.85\linewidth]{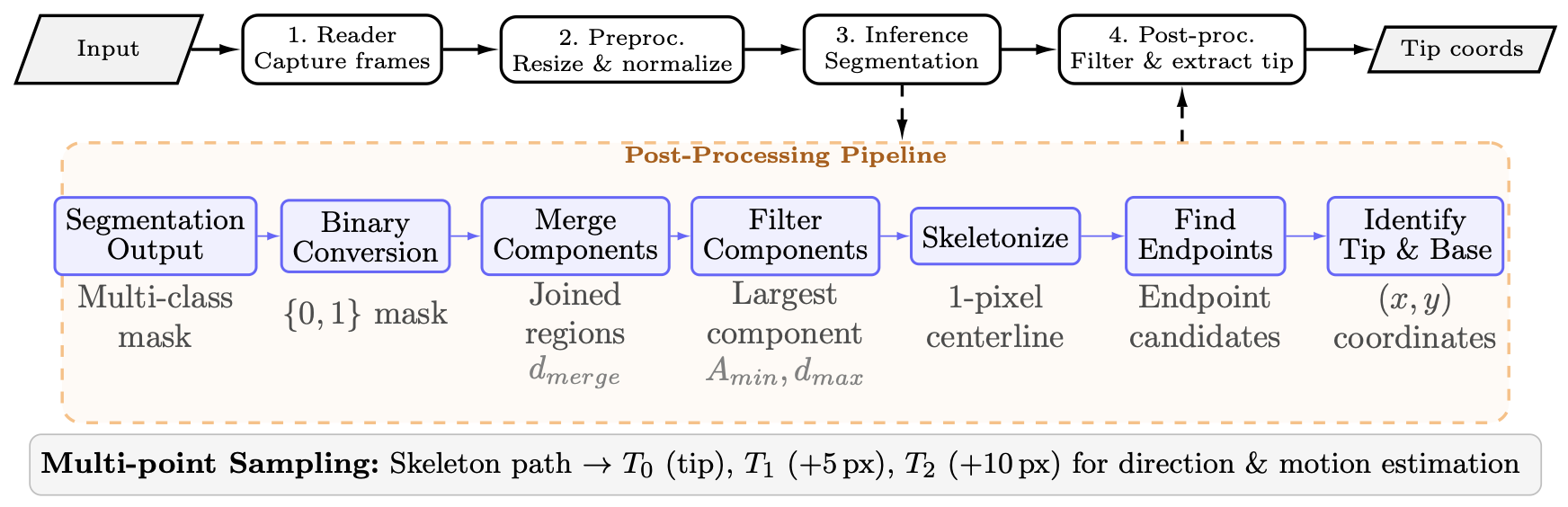}
            \caption{Four-stage tracking pipeline with expanded post-processing detail. The segmentation mask undergoes: binary conversion, component merging ($d_{merge}$), filtering by area and distance ($A_{min}$, $d_{max}$), skeletonization, and endpoint detection. Multi-point sampling ($T_0$, $T_1$, $T_2$) stabilizes direction and motion estimates.}
            \label{fig:tracking_pipeline}
        \end{figure}

        The device tracking algorithm operated as a four-stage asynchronous pipeline (Fig.~\ref{fig:tracking_pipeline}). Each stage ran on a dedicated thread to maximize throughput and minimize latency.
        
        \textbf{1. Reader:} Raw fluoroscopy or video frames were captured and enqueued for processing. This ensured that frame acquisition remained independent of model inference speed, avoiding dropped or skipped frames.
        
        \textbf{2. Preprocessor:} Each frame was resized, normalized, and converted to tensor format. This standardization ensured consistent input dimensions and intensity scaling across all images.
        
        \textbf{3. Inference:} Preprocessed frames were passed through a trained segmentation model to generate two or three-class masks. For large images ($\geq1024\times1024$\,pixel), a sliding-window approach with $512\times512$\,pixel patches and 256\,pixel overlap was used to balance segmentation quality and memory usage. Overlapping predictions were averaged to produce the final mask.
        
        \textbf{4. Post-processor:} The segmentation mask underwent several refinement steps: \textit{Component merging:} connected regions with overlapping bounding boxes were merged based on a defined pixel distance threshold. \textit{Principal structure selection:} the component with the largest area and closest extension to the inferior margin was selected as the primary structure. \textit{Artifact removal:} smaller or distant components (below area threshold or outside a fixed centroid distance from the principal tip) were discarded. \textit{Skeletonization:} a refined binary mask was thinned to a one-pixel-wide medial axis. \textit{Endpoint detection:} pixels with exactly one 8-connected neighbor were identified as endpoints. The base was defined as the smallest endpoint (minimum $y$-coordinate), and the tip as the endpoint farthest from the base in Euclidean distance. If fewer than two endpoints existed, contour-based tip detection was used. Hyperparameter selection for pixel distance threshold, area threshold, and bounding-box definitions used an iterative fine-tuning process for each dataset to determine the optimal values.
        
        Relying on a single tip coordinate can be unreliable due to motion blur or occlusion. To stabilize tracking and infer local direction, three points along the skeleton were sampled: the primary tip ($T_0$), and two downstream points located at arc-lengths of 5\,pixel ($T_1$) and 10\,pixel ($T_2$) from the tip. Tracking $T_0$, $T_1$, and $T_2$ is also necessary for the proposed system to be used in future autonomous endovascular navigation experiments, where many machine learning algorithms use multiple device coordinates starting from the tip as part of their input for each training step~\cite{Robertshaw2024,Robertshaw1_2025}.

    \subsection{Experimental design}
    
        The experimental design comprised two stages: training a range of models, and testing them in a variety of held out datasets to determine generalizability. All models were trained using Kaiming normal initialization with AdamW optimizer (lr=$1\!\times\!10^{-4}$, wd=$1\!\times\!10^{-5}$), cosine annealing ($T_{\text{max}}\!=\!100$--$150$, $\eta_{\text{min}}\!=\!10^{-6}$), batch size 16 (8 for SegFormer), and early stopping on validation loss (patience=10). Data augmentation included geometric (flips/rotations/transpose, $p\!=\!0.5$; elastic/grid distortion, $p\!=\!0.2$), intensity (brightness/contrast/noise, $p\!=\!0.3$; blur, $p\!=\!0.2$), and coarse dropout (10\% size, $p\!=\!0.2$). Inputs were normalized (mean=0.485, std=0.229) and trained with mixed-precision using Dice loss (smooth=$10^{-6}$) for 150-200 epochs. The TransU-Net employed 16 transformer layers and attention heads.

        \noindent\textbf{\textit{CathAction} dataset} This experiment compared binary (instrument vs.~background) and multi-class (catheter, guidewire, and background) segmentation tasks. Models were trained and validated using the CathAction dataset described in Section~\ref{sec:dataset}. This benchmark was used to verify correct implementation and integration of the segmentation backbones within the proposed tracking pipeline under controlled yet heterogeneous imaging conditions, before progressing to more challenging and clinically realistic \textit{in vivo} fluoroscopic data. Our U-Net implementation employed 16 base feature channels and was optimized using AdamW, whereas the original CathAction paper implemented U-Net with 64 base channels and stochastic gradient descent (SGD). This unified configuration was selected to satisfy real-time deployment constraints within the tracking pipeline, prioritizing latency and memory efficiency over reproducing architecture-specific peak segmentation performance, acknowledging that transformer-based models inherently require greater computational resources than pure CNNs. Therefore, performance comparisons to previously reported CathAction segmentation benchmarks should be interpreted in this context. Images were upsampled to $512 \times 512$.
        
        \noindent\textbf{\textit{In vitro offline RGB} dataset} Tracking accuracy was assessed using the mean absolute error along each image axis and in 2D Euclidean space: $\text{MAE}_x$, $\text{MAE}_y$ and $\text{MAE}_{(x,y)}$ (measured in mm), capturing the mean positional deviation between predicted and reference standard tip locations. Inference speed (IS) (measured in FPS) was also recorded.
        
        \noindent\textbf{\textit{In vitro live RGB} dataset} Real-time tracking was tested using a live RGB camera. Frames were downsampled to reduce inference latency. Models were evaluated qualitatively, as no reference standard was available.
        
        \noindent\textbf{\textit{In vivo fluoroscopic} dataset} To assess clinical applicability, models were evaluated on human fluoroscopic footage acquired during endovascular interventions. Reference standard masks were manually annotated and stratified by imaging complexity (Section~\ref{sec:dataset}). Tracking accuracy was assessed using $\text{MAE}_x$, $\text{MAE}_y$ and $\text{MAE}_{(x,y)}$ to provide a clinically interpretable measure of tracking under \textit{in vivo} fluoroscopy. No fine-tuning methods were used. IS was also recorded.
       
\section{Results}

    \subsection{\textit{CathAction}}
        
       Table~\ref{tab1} shows that SegFormer achieved the highest two-class performance (Dice: 0.809, IoU: 0.722). U-Net and U-Net+Transformer results were comparable, with Dice scores of 0.737 and 0.722, and IoUs of 0.653 and 0.640, respectively. SegFormer demonstrated the best recall (0.799) and $F_1$ score (0.809). For the three-class segmentation, overall performance decreased compared to the binary task. SegFormer achieved the best results across all metrics (Dice: 0.667, IoU: 0.557, $F_1$ score: 0.667). Previous benchmarks for three-class segementation on the CathAction dataset show Dice scores of 0.517 for U-Net, with the top highest score of 0.635 achieved using SegViT (different from our implementation of SegFormer)~\cite{Huang2024}.
        
       \begin{table}[h!]
            \footnotesize
            \centering
            \caption{Results across the \textit{CathAction} dataset.}
            \label{tab1}
            \begin{tabular}{lcccccc}     
                \toprule
                Algorithm & Dice & IoU & Precision & Recall & $F_1$ Score & IS \\
                \midrule
                \multicolumn{7}{c}{2-class (instrument, background)} \\
                \cmidrule(r){1-7}
                U-Net & 0.737 & 0.653 & 0.727 & 0.748 & 0.737 & \textbf{27.4} \\
                U-Net+Transformer & 0.722 & 0.640 & 0.702 & 0.747 & 0.722 & 24.5 \\
                SegFormer & \textbf{0.809} & \textbf{0.722} & \textbf{0.819} & \textbf{0.799} & \textbf{0.809} & 21.3 \\    
                \midrule
                \multicolumn{7}{c}{3-class (catheter, guidewire, background)} \\
                \midrule
                U-Net & 0.599 & 0.498 & 0.598 & 0.606 & 0.599 & \textbf{27.2} \\
                U-Net+Transformer & 0.587 & 0.488 & 0.573 & 0.616 & 0.587 & 25.0 \\
                SegFormer & \textbf{0.667} & \textbf{0.557} & \textbf{0.694} & \textbf{0.644} & \textbf{0.667} & 21.5 \\
                \bottomrule
            \end{tabular}
        \end{table}

    \subsection{\textit{In vitro offline RGB}}
    
\begin{figure}[h!]
            \centering
            \includegraphics[width=0.85\textwidth]{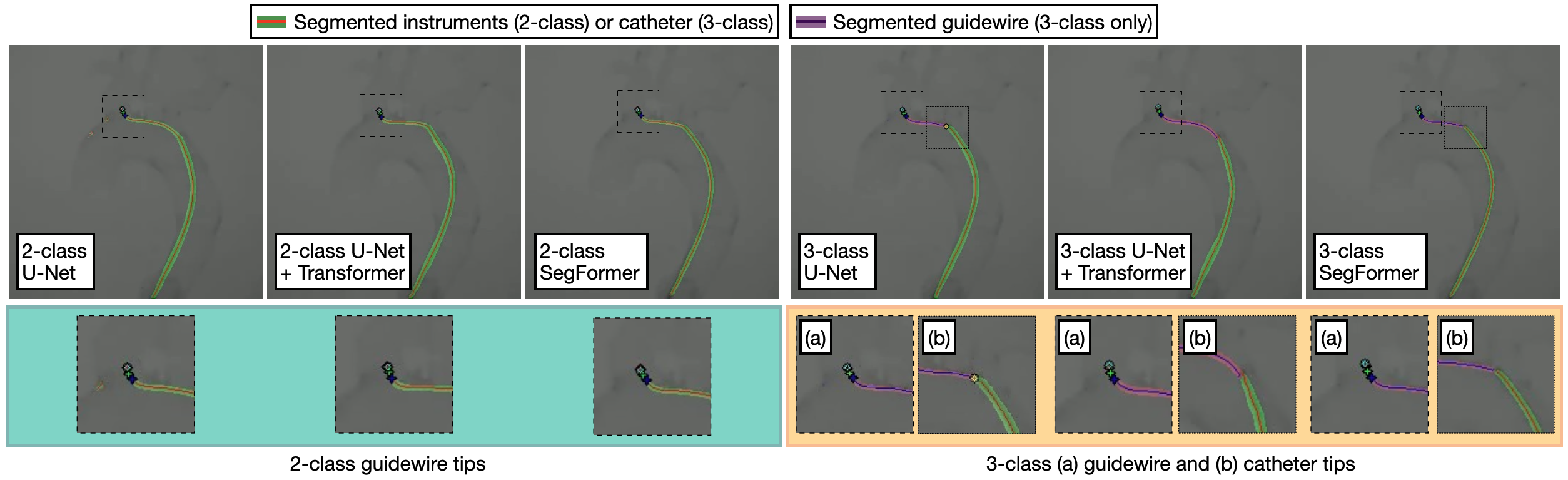}
            \caption{\textit{In vitro offline RGB} dataset results for binary and multi-class segmentation. Instrument tip annotations in first row have been enhanced for visualization purposes.}
            \label{fig:rgbv}
        \end{figure}

        For two-class segmentation, all three models achieved good localization of tips (Figure~\ref{fig:rgbv}). The three-class models suffer from misclassifications induced by skeleton branches and multi-pixel tip shifts near bifurcations, even after principal component isolation. Table~\ref{tab2} presents results on the \textit{in vitro offline RGB} dataset at $240 \times 320$ pixels resolution. All models achieved relatively low MAE values, with U-Net and U-Net+Transformer demonstrating comparable performance with $\text{MAE}_{(x, y)}$ at $0.52-0.53$\,mm for two-class, while SegFormer shows slightly lower errors with $\text{MAE}_{(x, y)}$ of $0.47$\,mm. For three-class, U-Net had the lowest $\text{MAE}_{(x, y)}$ (0.60\,mm) compared to U-Net+Transformer (0.66\,mm) and SegFormer (1.21\,mm). 

        \begin{table}[h]
            \footnotesize
            \caption{\textit{In vitro offline RGB} dataset results.}
            \label{tab2}
            \begin{tabular*}{\textwidth}{@{\extracolsep\fill}lcccccccc}
            \toprule%
            & \multicolumn{4}{@{}c@{}}{2-class} & \multicolumn{4}{@{}c@{}}{3-class} \\
            \cmidrule{2-5}\cmidrule{6-9}%
            Algorithm 
            & $\text{MAE}_{x}$ 
            & $\text{MAE}_{y}$ 
            & $\text{MAE}_{(x, y)}$ 
            & IS 
            & $\text{MAE}_{x}$ 
            & $\text{MAE}_{y}$ 
            & $\text{MAE}_{(x, y)}$ 
            & IS\\
            \cmidrule(r){1-9}
            U-Net  
            & 0.41 & 0.24 & 0.52 & 29.1
            & \textbf{0.42} & \textbf{0.33} & \textbf{0.60} & 29.3 \\
            U-Net+Transformer  
            & 0.43 & 0.24 & 0.53 & 29.7
            & 0.50 & 0.34 & 0.66 & 28.4 \\
            SegFormer  
            & \textbf{0.35} & \textbf{0.23} & \textbf{0.47} & \textbf{30.0}
            & 1.04 & 0.51 & 1.21 & \textbf{29.7} \\
            \bottomrule
            \end{tabular*}
        \end{table}

    \subsection{\textit{In vitro live RGB}}
            
\begin{figure}[h!]
            \centering
            \includegraphics[width=0.85\textwidth]{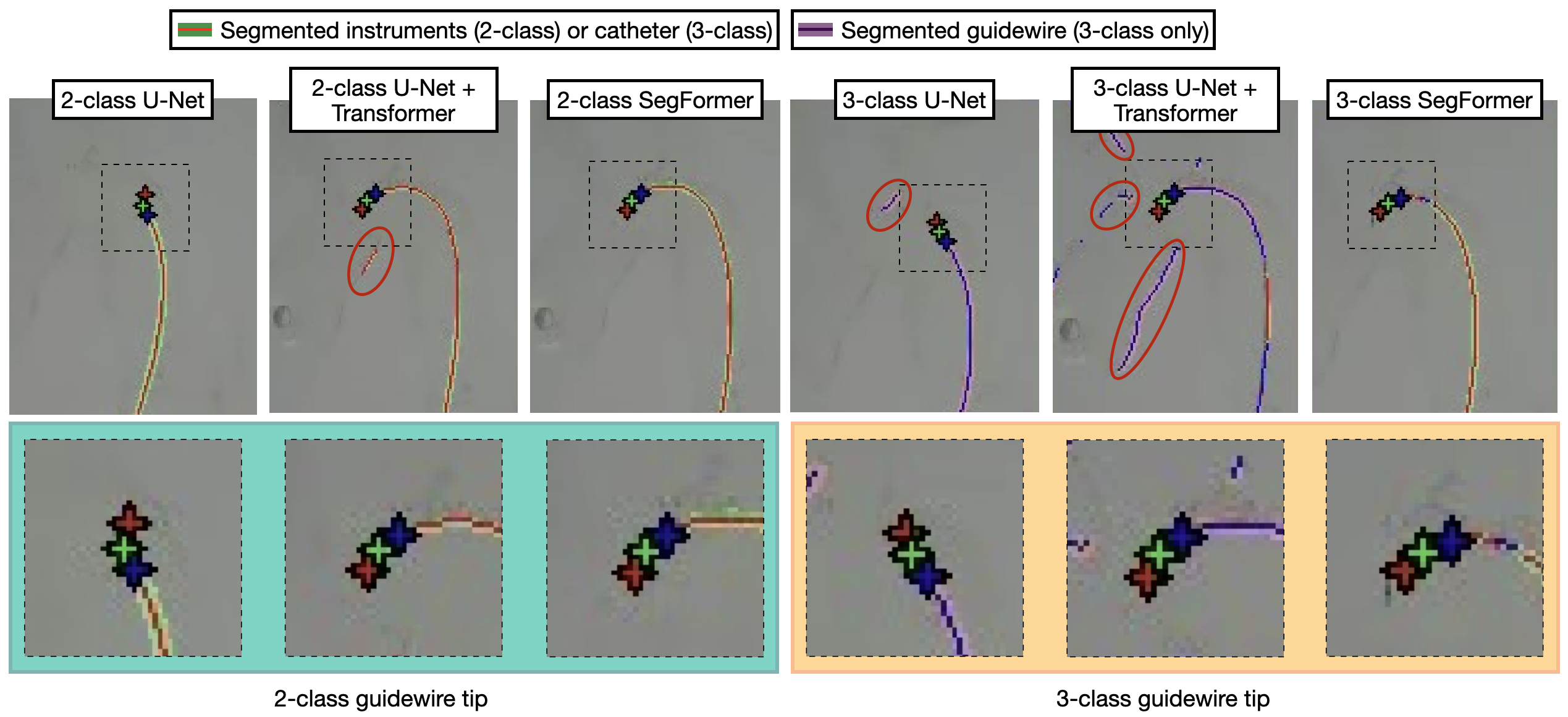}
            \caption{\textit{In vitro live RGB} dataset results. Instrument tip annotations in first row have been enhanced for visualization purposes. Misclassifications are circled in red.}
            \label{fig:camera}
        \end{figure}

        The representative images for the \textit{in vitro live RGB} dataset (Figure~\ref{fig:camera}) showed that the U-Net's tracking points deviated noticeably from the true apex, whereas U-Net+Transformer maintained tight alignment in both two and three-class modes. The transformer-based models demonstrated superior segmentation in complex backgrounds, while misclassification artifacts in three-class outputs persisted under live-camera conditions.
            
    \subsection{\textit{In vivo fluoroscopy}}
    
\begin{figure}[h!]
            \centering
            \includegraphics[width=0.85\textwidth]{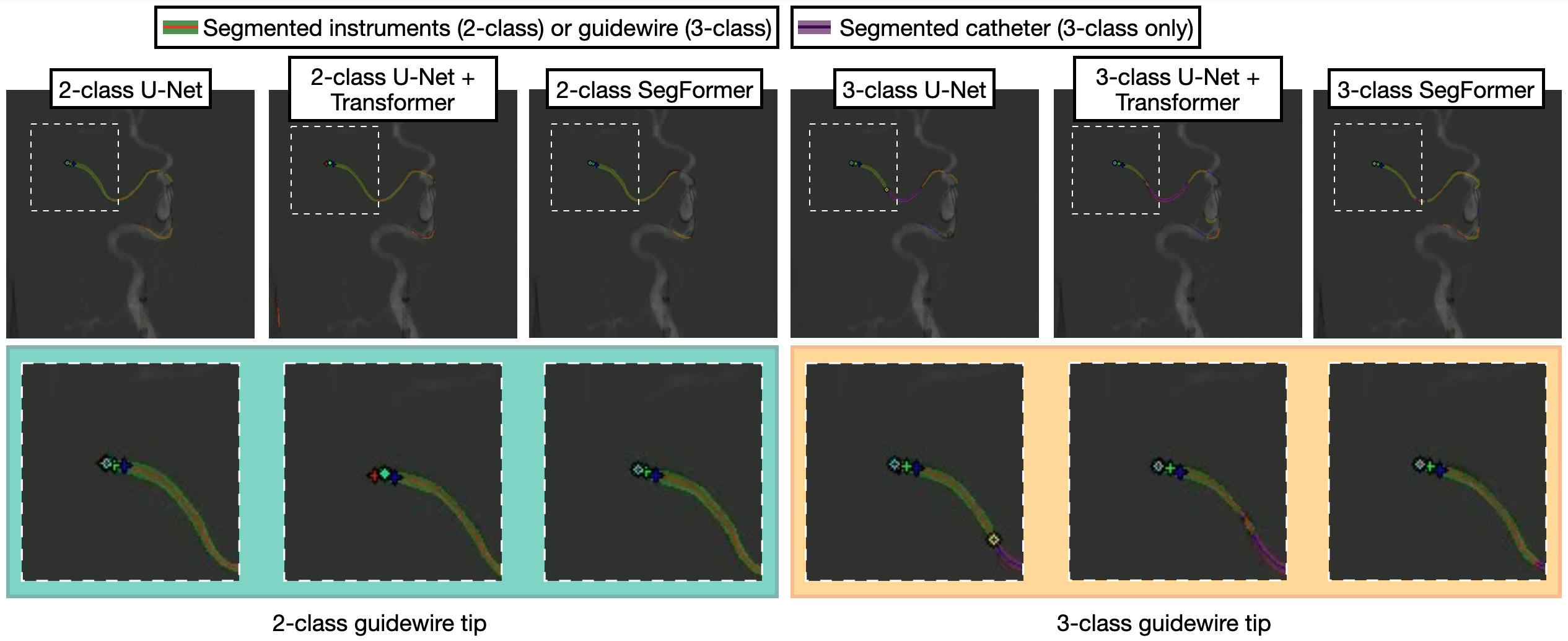}
            \caption{\textit{In vivo fluoroscopic} dataset for G1 (high complexity) example. Instrument tip annotations in first row have been enhanced for visualization purposes.}
            \label{fig:fluoro}
        \end{figure}

        Representative images of the segmentation task for G1: high-complexity are shown in Figure~\ref{fig:fluoro}. In contrast to the phantom test results, the Segformer model with U-Net decoder exhibited superior robustness when processing complex vascular structures. Most notably, the Segformer architecture generated minimal artifacts in the presence of complex anatomical backgrounds, eliminating the need for bounding box-based background filtering typically required for accurate tip tracking. 
        
        Tip tracking performance on the \textit{in vivo fluoroscopic} datasets was evaluated via MAE against ground truth masks of the $x, y$ and $(x, y)$ coordinates, which is presented in Table~\ref{tab3}. Higher MAE values were observed in fluoroscopic data relative to phantom data, due to greater visual variability and complexity. SegFormer achieved the lowest $\text{MAE}_{(x,y)}$ in G1 and G2 two-class datasets.

        \begin{table}[h!]
            \footnotesize
            \centering
            \caption{\textit{In vivo} fluoroscopy segmentation results.}
            \label{tab3}
            \begin{tabular*}{\textwidth}{@{\extracolsep\fill}lcccccccc}
                \toprule
                \textbf{Algorithm} 
                & \multicolumn{4}{c}{\textbf{2-Class}} 
                & \multicolumn{4}{c}{\textbf{3-Class}} \\
                \cmidrule(lr){2-5} \cmidrule(lr){6-9}
                & $\text{MAE}_{x}$ & $\text{MAE}_{y}$ & $\text{MAE}_{(x,y)}$ & IS 
                & $\text{MAE}_{x}$ & $\text{MAE}_{y}$ & $\text{MAE}_{(x,y)}$ & IS \\
                \midrule
        
                \multicolumn{9}{c}{\textbf{G1: High Complexity}} \\
                \midrule
                U-Net & 2.24 & 11.86 & 12.18 & \textbf{61.7} & 5.29 & 27.73 & 29.02 & \textbf{61.8} \\
                U-Net+Transformer & 1.47 & 7.96 & 8.17 & 53.9 & \textbf{2.58} & \textbf{9.52} & \textbf{10.35} & 57.7 \\
                SegFormer & \textbf{2.03} & \textbf{6.96} & \textbf{7.58} & 36.1 & 1.40 & 24.07 & 24.19 & 34.6 \\
        
                \midrule
                \multicolumn{9}{c}{\textbf{G2: Moderate Complexity}} \\
                \midrule
                U-Net & 3.10 & 2.82 & 4.60 & 32.0 & 4.46 & 5.41 & 7.74 & 25.0 \\
                U-Net+Transformer & 4.28 & 3.72 & 6.20 & \textbf{35.6} & 4.09 & 3.01 & 5.59 & \textbf{33.9} \\
                SegFormer & \textbf{2.98} & \textbf{2.91} & \textbf{4.44} & 30.5 & \textbf{3.44} & \textbf{3.47} & \textbf{5.19} & 30.2 \\
        
                \midrule
                \multicolumn{9}{c}{\textbf{G3: Low Complexity}} \\
                \midrule
                U-Net & \textbf{2.94} & \textbf{2.48} & \textbf{4.28} & 30.6 & 3.88 & 4.04 & 6.08 & \textbf{43.5} \\
                U-Net+Transformer & 4.09 & 4.56 & 6.78 & \textbf{39.0} & 4.26 & 3.79 & 6.06 & 42.4 \\
                SegFormer & 2.71 & 2.92 & 4.37 & 30.9 & \textbf{3.70} & \textbf{2.51} & \textbf{4.79} & 30.1 \\
        
                \bottomrule
            \end{tabular*}
        \end{table}

\section{Discussion}

    This work presents a fully integrated, real-time catheter tracking pipeline capable of robust performance across \textit{in vitro} phantom, \textit{in vivo} animal, and \textit{in vivo} clinical fluoroscopy data. Results demonstrate that the proposed multi-threaded tracking framework maintains stable performance under challenging imaging conditions, confirming its potential for translation into AI-assisted endovascular navigation. On clinical fluoroscopic data, SegFormer appeared to exhibit the most stable and accurate tip localization, with the lowest MAE across datasets, demonstrating its ability to generalize to complex, variable anatomical backgrounds. 
    
    Our segmentation results demonstrate that U-Net and U-Net+Transformer architectures perform well on the CathAction phantom dataset. Using CathAction as a controlled benchmark allowed us to verify correct implementation of the segmentation backbones within the proposed tracking pipeline and to contextualize performance relative to prior work. Compared with the original CathAction benchmarks~\cite{Huang2024}, we observed consistent improvements across segmentation metrics when evaluated within our experimental setup. U-Net generally achieved higher performance scores, whereas U-Net+Transformer maintained slightly higher recall, suggesting that self-attention mechanisms aid in recovering additional true positives at the expense of some false positives. These findings highlight a trade-off between overall segmentation accuracy and recall, particularly in challenging or low-contrast regions. For three-class segmentation, metrics decreased relative to two-class segmentation due to increased class complexity and low-contrast structures. It is plausible that the global context modeling of transformer-based architectures can help recover subtle guidewire segments, but additional class complexity may challenge precise segmentation and tip localization.

    Tip-tracking evaluation revealed consistent performance across models on \textit{in vitro} phantom hold-out datasets (\textit{in vitro offline RGB} and \textit{in vitro live RGB}). Although transformer-based models introduced global context information, they did not automatically improve tip localization in all cases, suggesting that further regularization or data diversity may be required. \textit{In vitro offline RGB} testing highlighted differences between static phantom benchmarks and dynamic, real-world environments. Two-class models consistently outperformed three-class variants in both speed and tip-localization accuracy, while three-class models suffered from misclassification-induced skeleton branches and multi-pixel tip shifts near bifurcations, suggesting a trade-off between anatomical detail and real-time tip-tracking performance. The benefits of attention mechanisms in transformer-based models appeared to be more apparent in complex backgrounds, supporting their potential value. \textit{In vitro live RGB} results potentially supported these observations. While U-Net tip points appeared to deviate under suboptimal imaging conditions, transformer-based models looked to maintain tighter alignment, demonstrating increased robustness to noise and variability. Misclassification artifacts seemed to persist in three-class outputs, reinforcing that richer anatomical labeling may potentially require more sophisticated architectures for precise tip-localization. 

    Overall, it is plausible that two-class segmentation provides the most reliable trade-off between speed, stability, and tip-localization accuracy, while 3-class segmentation, despite slightly lower accuracy and increased computational cost, has demonstrated advantages in autonomous RL navigation by enabling more effective simultaneous use of both catheter and guidewire devices~\cite{Robertshaw2024}. As research progresses from \textit{in silico} and \textit{in vitro} studies towards clinical translation, this dual-tracking capability may be key in allowing autonomous agents to navigate diverse patient vasculatures, providing richer input to RL agents and supporting more sophisticated real-time decision-making. These insights can inform the design of future real-time, clinically deployable tracking algorithms, particularly when deploying models on hardware-constrained or clinically variable imaging environments. Our study uses solely the CathAction phantom dataset for training, with no \textit{in vivo} fine-tuning. On \textit{in vivo} fluoroscopy, the observed MAE values (approximately $4–7$\,mm depending on complexity) remain above the $<1$\,mm error reported in prior segmentation-focused studies~\cite{Ambrosini2017}. However, these thresholds typically reflect segmentation-level evaluation rather than full downstream tracking under occlusion, motion, and anatomical variability. In contrast, our MAE reflects the complete tracking pipeline (segmentation, skeletonization, endpoint detection, and multi-point sampling) applied to heterogeneous clinical fluoroscopy without domain adaptation such as fine tuning. 

    The gap between phantom ($<1$\,mm) and \textit{in vivo} performance may reflect: (1) domain shift in contrast, noise, and motion; (2) partial occlusion and overlapping instruments; (3) inter-frame motion blur; and (4) absence of \textit{in vivo} training data. This study should therefore be interpreted as demonstrating feasibility and cross-domain generalization rather than clinical-grade precision. Incorporating \textit{in vivo} fluoroscopic training data and domain adaptation strategies may reduce tracking error toward clinically actionable thresholds.

    While the proposed tracking framework demonstrated strong experimental performance and shows potential for clinical translation, limitations remain. Tip annotations were performed by non-clinicians, introducing potential labeling bias and uncertainty, while data labeling for hold-out data was restricted to a small number of test videos ($\approx 15$\,s each), whereas real procedures may last several hours, limiting the system’s clinical representativeness. The multi-threaded processing pipeline, though optimized for real-time, may experience minor synchronization delays during high-frequency operations, and long-term runtime stability under continuous clinical use remains untested. Future work will aim to develop high-fidelity phantom-based evaluation environments for improved generalization, integrate expert-annotated \textit{in vivo} fluoroscopic datasets from diverse anatomies to reduce domain gaps and improve reliability, while performing extended-duration trials across multiple full-length MT interventions to assess performance, stability, and synchronization over time. It should also look to examine multi-point tracking under changing imaging conditions, such as zooming operations, patient breathing, changes in viewpoint, or other non-planar motions. This would allow for further evidence that the proposed system is able to be be used in current autonomous navigation algorithms for the continued development of autonomous tele-operated robotic platforms~\cite{Robertshaw2024,Robertshaw1_2025}.

\section{Conclusion}

    This study proposed and evaluated a deep learning–based catheter/guidewire tip-tracking framework using video and live-camera data of a phantom, and clinical fluoroscopic data. Among the tested architectures, the transformer-based SegFormer achieved the lowest localization error and demonstrating improved robustness relative to convolutional baselines under anatomical complexity, though clinical-grade sub-millimeter precision was not achieved in \textit{in vivo} fluoroscopy. To further improve its translational potential, further work is required to address domain gaps, assess robustness and stability under real interventional conditions with full procedure duration. By improving model generalization, real-time performance, and temporal stability, and integrating the tracking pipeline with RL-based navigation, we can advance toward clinically deployable, real-time device tracking systems for autonomous endovascular interventions.

\bibliographystyle{unsrt}  
\bibliography{references}  

@article{Jadhav2021,
   author = {Ashutosh P. Jadhav and Shashvat M. Desai and Tudor G. Jovin},
   doi = {10.1212/WNL.0000000000012801},
   issn = {1526632X},
   issue = {20},
   journal = {Neurology},
   pmid = {34785611},
   publisher = {Lippincott Williams and Wilkins},
   title = {Indications for Mechanical Thrombectomy for Acute Ischemic Stroke: Current Guidelines and Beyond},
   volume = {97},
   year = {2021}
}

@article{Saver2016,
   author = {Jeffrey L Saver and Mayank Goyal and Aad van der Lugt and Bijoy K Menon, Charles B L M Majoie and Diederik W Dippel and Bruce C Campbell and Raul G Nogueira and Andrew M Demchuk and Alejandro Tomasello and Pere Cardona and Thomas G Devlin and Donald F Frei and Richard du Mesnil de Rochemont and Olvert A Berkhemer and Tudor G Jovin and Adnan H Siddiqui and Wim H van Zwam and Stephen M Davis and Carlos Castaño and Biggya L Sapkota and Puck S Fransen and Carlos Molina and Robert J van Oostenbrugge and Ángel Chamorro and Hester Lingsma and Frank L Silver and Geoffrey A Donnan and Ashfaq Shuaib and Scott Brown and Bruce Stouch and Peter J Mitchell and Antoni Davalos and Yvo B W E M Roos and Michael D Hill and {HERMES Collaborators}},
   doi = {10.1001/jama.2016.13647},
   issn = {15383598},
   issue = {12},
   journal = {JAMA - Journal of the American Medical Association},
   month = {9},
   pages = {1279-1288},
   pmid = {27673305},
   publisher = {American Medical Association},
   title = {Time to treatment with endovascular thrombectomy and outcomes from ischemic stroke: Ameta-analysis},
   volume = {316},
   year = {2016}
}

@misc{SSNAP2024,
   author = {{Sentinel Stroke National Audit Programme}},
   title = {SSNAP Annual Report 2024},
   url = {https://www.hqip.org.uk/resource/ssnap-nov24/},
   year = {2024},
   month = {11},
   day = {14},
   note = {Accessed 5 March 2025},
   howpublished = {Healthcare Quality Improvement Partnership},
}

@article{Robertshaw2023,
   author = {Harry Robertshaw and Lennart Karstensen and Benjamin Jackson and Hadi Sadati and Kawal Rhode and Sebastien Ourselin and Alejandro Granados and Thomas C. Booth},
   doi = {10.3389/fnhum.2023.1239374},
   issn = {1662-5161},
   journal = {Frontiers in Human Neuroscience},
   title = {Artificial intelligence in the autonomous navigation of endovascular interventions: a systematic review},
   year = {2023}
}

@article{Berkhemer2015,
   author = {O. A. Berkhemer and P. S. S. Fransen and D. Beumer and L. A. van den Berg and H. F. Lingsma and A. J. Yoo and W. J. Schonewille and J. A. Vos and P. J. Nederkoorn and M. J. H. Wermer and M. A. A. van Walderveen and J. Staals and J. Hofmeijer and J. A. van Oostayen and G. J. Lycklama a Nijeholt and J. Boiten and P. A. Brouwer and B. J. Emmer and S. F. de Bruijn and L. C. van Dijk and L. J. Kappelle and R. H. Lo and E. J. van Dijk and J. de Vries and P. L. M. de Kort and W. J. J. van Rooij and J. S. P. van den Berg and B. A. A. M. van Hasselt and L. A. M. Aerden and R. J. Dallinga and M. C. Visser and J. C. J. Bot and P. C. Vroomen and O. Eshghi and T. H. C. M. L. Schreuder and R. J. J. Heijboer and K. Keizer and A. V. Tielbeek and H. M. den Hertog and D. G. Gerrits and R. M. van den Berg-Vos and G. B. Karas and E. W. Steyerberg and H. Z. Flach and H. A. Marquering and M. E. S. Sprengers and S. F. M. Jenniskens and L. F. M. Beenen and R. van den Berg and P. J. Koudstaal and W. H. van Zwam and Y. B. W. E. M. Roos and A. van der Lugt and R. J. van Oostenbrugge and C. B. L. M. Majoie and D. W. J. Dippel},
   doi = {10.1056/nejmoa1411587},
   issn = {0028-4793},
   issue = {1},
   journal = {New England Journal of Medicine},
   month = {1},
   pages = {11-20},
   pmid = {25517348},
   publisher = {Massachusetts Medical Society},
   title = {A Randomized Trial of Intraarterial Treatment for Acute Ischemic Stroke},
   volume = {372},
   year = {2015}
}

@article{Klein2009,
   author = {Lloyd W Klein and Donald L Miller and Stephen Balter and Warren Laskey and David Haines and Alexander Norbash and Matthew A. Mauro and James A. Goldstein},
   issue = {2},
   journal = {Society of Interventional Radiology},
   month = {2},
   pages = {538-544},
   title = {Occupational Health Hazards in the Interventional Laboratory: Time for a Safer Environment},
   volume = {250},
   year = {2009}
}

@article{Madder2017,
   author = {Ryan D. Madder and Stacie VanOosterhout and Abbey Mulder and and Matthew Elmore and Jessica Campbell and Andrew Borgman and Jessica Parker and David Wohns},
   doi = {10.1016/j.carrev.2016.12.011},
   issn = {18780938},
   issue = {3},
   journal = {Cardiovascular Revascularization Medicine},
   month = {4},
   pages = {190-196},
   pmid = {28041859},
   publisher = {Elsevier Inc.},
   title = {Impact of robotics and a suspended lead suit on physician radiation exposure during percutaneous coronary intervention},
   volume = {18},
   year = {2017}
}

@article{Crinnion2022,
   author = {William Crinnion and Ben Jackson and Avnish Sood and Jeremy Lynch and Christos Bergeles and Hongbin Liu and Kawal Rhode and Vitor Mendes Pereira and Thomas C Booth},
   doi = {10.1136/neurintsurg-2021-018096},
   issn = {17598486},
   issue = {6},
   journal = {Journal of neurointerventional surgery},
   month = {6},
   pages = {539-545},
   pmid = {34799439},
   publisher = {NLM (Medline)},
   title = {Robotics in neurointerventional surgery: a systematic review of the literature},
   volume = {14},
   year = {2022}
}

@article{Mofatteh2021,
   author = {Mohammad Mofatteh},
   doi = {10.3934/Neuroscience.2021025},
   issn = {23737972},
   issue = {4},
   journal = {AIMS Neuroscience},
   pages = {477-495},
   publisher = {AIMS Press},
   title = {Neurosurgery and artificial intelligence},
   volume = {8},
   year = {2021}
}

@article{Robertshaw2024,
   author = {Harry Robertshaw and Lennart Karstensen and Benjamin Jackson and Alejandro Granados and Thomas C. Booth},
   doi = {10.1007/s11548-024-03208-w},
   issn = {1861-6429},
   journal = {Int J CARS},
   month = {6},
   title = {Autonomous navigation of catheters and guidewires in mechanical thrombectomy using inverse reinforcement learning},
   year = {2024}
}

@article{Robertshaw1_2025,
   author = {Harry Robertshaw and Benjamin Jackson and Jiaheng Wang and Hadi Sadati and Lennart Karstensen and Alejandro Granados and Thomas C. Booth},
   doi = {10.1007/s11548-025-03339-8},
   issn = {18616429},
   journal = {Int J CARS},
   publisher = {Springer Science and Business Media Deutschland GmbH},
   title = {Reinforcement learning for safe autonomous two-device navigation of cerebral vessels in mechanical thrombectomy},
   year = {2025}
}

@article{Jianu2024,
   author = {Tudor Jianu and Baoru Huang and Minh Nhat Vu and Mohamed E. M. K. Abdelaziz and Sebastiano Fichera and Chun-Yi Lee and Pierre Berthet-Rayne and Ferdinando Rodriguez y Baena and Anh Nguyen},
   doi = {10.1109/TMRB.2024.3421256},
   issn = {25763202},
   issue = {3},
   journal = {IEEE Transactions on Medical Robotics and Bionics},
   publisher = {Institute of Electrical and Electronics Engineers Inc.},
   title = {CathSim: An Open-Source Simulator for Endovascular Intervention},
   year = {2024}
}

@inproceedings{Ambrosini2017,
   title = "Fully Automatic and Real-Time Catheter Segmentation in X-Ray Fluoroscopy",
   author = "Pierre Ambrosini and D Ruijters and Wiro Niessen and Adriaan Moelker and van Walsum, Theo",
   year = "2017",
   doi = "10.1007/978-3-319-66185-8\_65",
   language = "English",
   isbn = "9783319661858",
   booktitle = "Medical Image Computing and Computer-Assisted Intervention 2017",
}

@INPROCEEDINGS{Nguyen2020,
   author={Nguyen, Anh and Kundrat, Dennis and Dagnino, Giulio and Wenqiang Chi and Mohamed E. M. K. Abdelaziz and Yao Guo and YingLiang Ma and Trevor M. Y. Kwok and Celia Riga and Guang-Zhong Yang},
   booktitle={2020 IEEE International Conference on Robotics and Automation (ICRA)}, 
   title={End-to-End Real-time Catheter Segmentation with Optical Flow-Guided Warping during Endovascular Intervention}, 
   year={2020},
   pages={9967-9973},
   doi={10.1109/ICRA40945.2020.9197307}
}

@inproceedings{Vlontzos2018,
   author={Athanasios Vlontzos and Krystian Mikolajczyk},
   title={Deep Segmentation and Registration in X-Ray Angiography Video},
   booktitle={British Machine Vision Conference 2018},
   pages={267},
   year={2018},
}

@inproceedings{Demoustier2023,
   author = {Demoustier, Marc and Zhang, Yue and Murthy, Venkatesh and Ghesu, Florin and Comaniciu, Dorin},
   year = {2023},
   pages = {679-688},
   title = {ConTrack: Contextual Transformer for Device Tracking in X-Ray},
   isbn = {978-3-031-43995-7},
   doi = {10.1007/978-3-031-43996-4_65},
   booktitle = "Medical Image Computing and Computer-Assisted Intervention 2013",
}

@techreport{Huang2024,
   title = "CathAction: A Benchmark for Endovascular Intervention Understanding",
   author = "Baoru Huang and Tuan Vo and Chayun Kongtongvattana and Giulio Dagnino and Dennis Kundrat and Wenqiang Chi and Mohamed Abdelaziz and Trevor Kwok and Tudor Jianu and Tuong Do and Hieu Le and Minh Nguyen and Hoan Nguyen and Erman Tjiputra and Quang Tran and Jianyang Xie and Yanda Meng and Binod Bhattarai and Zhaorui Tan and Hongbin Liu and Hong Seng Gan and Wei Wang and Xi Yang and Qiufeng Wang and Jionglong Su and Kaizhu Huang and Angelos Stefanidis and Min Guo and Bo Du and Rong Tao and Minh Vu and Guoyan Zheng and Yalin Zheng and Francisco Vasconcelos and Danail Stoyanov and Daniel Elson and Ferdinando Rodriguez y Baena and Anh Nguyen",
   year={2024},
   eprint={2408.13126},
   archivePrefix={arXiv},
   primaryClass={cs.CV},
   url={https://arxiv.org/abs/2408.13126}, 
}

@InProceedings{Ronneberger2015,
   author="Ronneberger, Olaf and Fischer, Philipp and Brox, Thomas",
   title="U-Net: Convolutional Networks for Biomedical Image Segmentation",
   booktitle="Medical Image Computing and Computer-Assisted Intervention 2015",
   year="2015",
   pages="234--241",
   isbn="978-3-319-24574-4"
}

@article{Chen2024,
   author = {Jieneng Chen and Jieru Mei and Xianhang Li and Yongyi Lu and Qihang Yu andQingyue Wei and Xiangde Luo and Yutong Xie and Ehsan Adeli and Yan Wang and Matthew P. Lungren and Shaoting Zhang and Lei Xing and Le Lu and Alan Yuille and Yuyin Zhou},
   doi = {10.1016/j.media.2024.103280},
   issn = {13618423},
   journal = {Medical Image Analysis},
   month = {10},
   pmid = {39096845},
   publisher = {Elsevier B.V.},
   title = {TransUNet: Rethinking the U-Net architecture design for medical image segmentation through the lens of transformers},
   volume = {97},
   year = {2024}
}

@InProceedings{Cao2023,
   author="Cao, Hu and Wang, Yueyue and Chen, Joy and Dongsheng Jiang and Xiaopeng Zhang and Qi Tian and Manning Wang",
   title="Swin-Unet: Unet-Like Pure Transformer for Medical Image Segmentation",
   booktitle="Computer Vision -- ECCV 2022 Workshops",
   year="2023",
   pages="205--218",
   isbn="978-3-031-25066-8"
}

@InProceedings{Zhang2022,
   author = {Zhang, Bowen and Tian, Zhi and Tang, Quan and Xiangxiang Chu and Xiaolin Wei and Chunhua Shen and Yifan Liu},
   year = {2022},
   month = {10},
   pages = {},
   title = {SegViT: Semantic Segmentation with Plain Vision Transformers},
   doi = {10.48550/arXiv.2210.05844},
   booktitle = {36th Conference on Neural Information Processing Systems}
}

@article{xie2021segformer,
  title={SegFormer: Simple and efficient design for semantic segmentation with transformers},
  author={Xie, Enze and Wang, Wenhai and Yu, Zhiding and Anima Anandkumar and Jose M. Alvarez and Ping Luo},
  journal={Advances in neural information processing systems},
  volume={34},
  pages={12077--12090},
  year={2021}
}

@article{GHERARDINI2020105420,
title = {Catheter segmentation in X-ray fluoroscopy using synthetic data and transfer learning with light U-nets},
journal = {Computer Methods and Programs in Biomedicine},
volume = {192},
pages = {105420},
year = {2020},
issn = {0169-2607},
doi = {https://doi.org/10.1016/j.cmpb.2020.105420},
author = {Marta Gherardini and Evangelos Mazomenos and Arianna Menciassi and Danail Stoyanov},
}

\end{document}